\documentclass[11pt]{article}

\usepackage[final]{acl}

\usepackage{times}
\usepackage{latexsym}

\usepackage[T1]{fontenc}

\usepackage[utf8]{inputenc}

\usepackage{microtype}

\usepackage{inconsolata}

\usepackage{graphicx}

\usepackage{quoting}
\usepackage{booktabs}
\usepackage{multirow}
\usepackage{makecell}

\usepackage{nicematrix}
\usepackage{tabularray}
\usepackage{pifont}
\usepackage{tikz}
\newcommand{\tr}[1]{\textcolor{red}{#1}}
\usepackage{CJKutf8}
\usepackage{amsmath}
\usepackage{enumitem}

\usepackage{tcolorbox}
\usepackage{listings}
\usepackage{xcolor}
\usepackage{fancyvrb}
\definecolor{lightblue}{RGB}{46,110,187}
\definecolor{darkred}{RGB}{150,38,31}
\definecolor{darkgreen}{rgb}{0.0, 0.5, 0.0}
\definecolor{blue}{HTML}{3572EF}
\newcommand{\tg}[1]{\textcolor{darkgreen}{#1}}
\newcommand{\tb}[1]{\textcolor{blue}{#1}}
\usepackage{lipsum}

\usepackage{arydshln}
\usepackage{subcaption} 

\title{MetFuse: Figurative Fusion between Metonymy and Metaphor}

\author{Saptarshi Ghosh \and Tianyu Jiang \\
  University of Cincinnati \\
  \texttt{ghosh2si@mail.uc.edu, tianyu.jiang@uc.edu} \\}

\begin{document}
\maketitle

\begin{abstract}
Metonymy and metaphor often co-occur in natural language, yet computational work has studied them largely in isolation. We introduce a framework that transforms a literal sentence into
three figurative variants: metonymic, metaphoric, and hybrid. Using this framework, we construct \textbf{MetFuse},\footnote{\url{https://github.com/cincynlp/MetFuse}} the first dedicated dataset of figurative fusion between metonymy and metaphor, containing 1,000 human-verified meaning-aligned quadruplets totaling 4,000 sentences. Extrinsic experiments on eight existing benchmarks show that augmenting training data with MetFuse consistently improves both metonymy and metaphor classification, with hybrid examples yielding the largest gains on metonymy tasks. Using this dataset, we also analyze how the presence of one figurative type impacts another. Our findings show that both human annotators and large language models better identify metonymy in hybrid sentences than in metonymy-only sentences, demonstrating that the presence of a metaphor makes a metonymic noun more explicit.
\end{abstract}

\section{Introduction}

Metonymy and metaphor are two fundamental linguistic phenomena in figurative language that involve concept mapping~\citep{radden}. While both entail a shift in meaning, they operate through distinct mechanisms. Metonymy primarily occurs through a change in meaning of the noun. Metaphors, by contrast, are more varied in form. Verbal metaphors are a prominent subtype, where the figurative shift arises through the verb. The crucial difference between metonymy and metaphor is that in metaphoric mapping, two discrete domains are involved, whereas mapping in metonymy occurs within a single domain~\citep{gossens, lakoff1980}. 

\begin{figure}[t]
    \centering
    \includegraphics[width=0.98\linewidth]{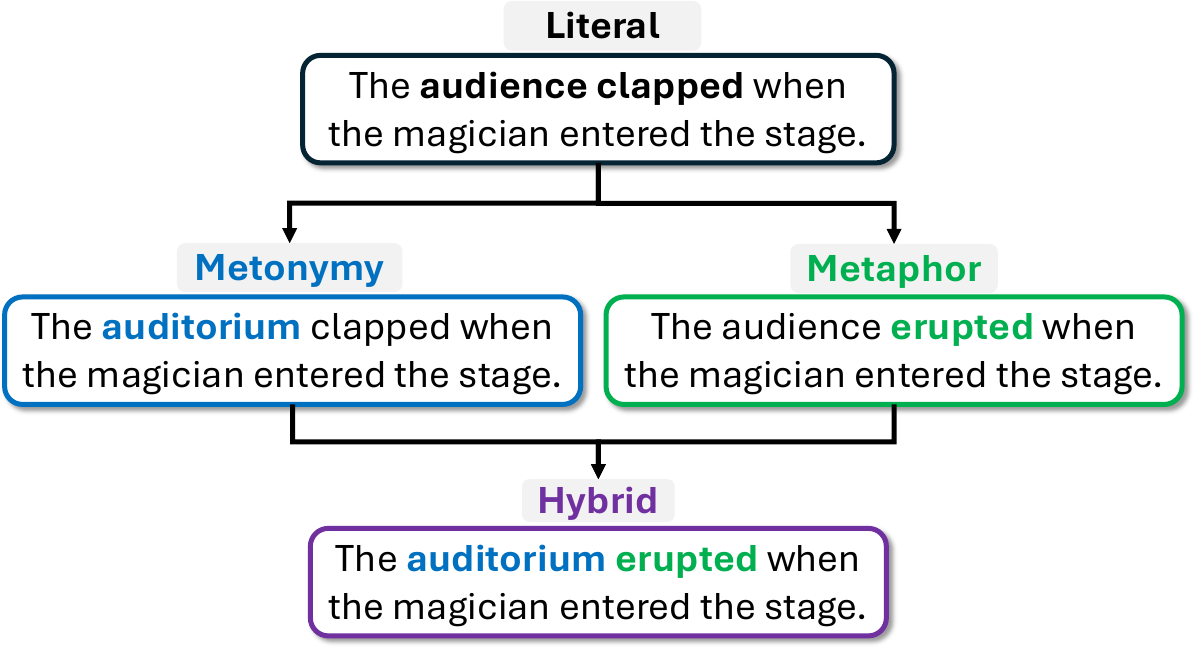}
    \caption{Illustration of generating metonymy, metaphor and hybrid sentences from literal expressions.}
    \label{fig:intro_figure}
\end{figure}

For instance, in the metonymic sentence ``\textit{The \textbf{stadium} celebrated joyfully},''  the noun \textit{stadium} (a location) refers to the fans in the stadium. The mapping stays within the same domain because both the location and the people associated with it belong to the broader conceptual domain of real-world entities tied to a place. Now looking at the metaphor ``\textit{The fans \textbf{erupted} with joy},'' the verb \textit{erupted} highlights the intense emotional reaction of the fans through the domain of physical explosion, making it a cross-domain mapping. Hence, metonymy preserves domain continuity, while metaphor requires a conceptual leap across domains---making it structurally and cognitively more distinct~\citep{lakoff1980}. Interestingly, a metonymic noun and metaphoric verb can co-exist in a sentence particularly in creative writing, resulting in figurative fusion like ``\textit{The \textbf{stadium erupted} with joy}.'' Such constructions are powerful because they capture multiple layers of meaning into a single expression, enriching the expressive potential of the language, enabling writers to create vivid, memorable scenes that resonate with readers at both conceptual and emotional level~\citep{lakoff1980, turner}. By modeling these combinations, we can push LLMs toward generating more nuanced, stylistically rich text, which is crucial for applications in narrative generation and other creative domains.

Prior linguistic works have studied metonymy and metaphor separately, primarily focusing on their differences in terms of their linguistic and cognitive nature~\citep{lakoff1980, radden}. There are also some linguistic works, namely that of \citet{gossens} and \citet{barcelona}, that study metonymy and metaphor together, theorizing their interactions in context. However, these theoretical works were constrained by the lack of resources~\citep{gossens}. To our knowledge, very few works have explored metonymy and metaphor jointly in the context of modern NLP and LLMs. 

Filling in these research gaps, we explore the fusion of metonymy and metaphor. We introduce a literal-to-figurative transformation framework that takes a literal sentence as input, and produces three outputs: a metonymic, a metaphoric and a hybrid (containing both metonymy and metaphor) variant of the sentence, as depicted in Figure~\ref{fig:intro_figure}. Our approach significantly outperforms baselines. Human annotation statistics show that 78\% of the generated sentences using our framework were judged as valid realizations of the intended figurative category, compared to only 53\% when using a straightforward prompt.

Leveraging this framework, we introduce \textbf{MetFuse}, the first dataset of figurative fusion between metonymy and metaphor, containing 1,000 quadruplets, each comprising a literal sentence paired with its metonymic, metaphoric and hybrid counterparts, yielding 4,000 sentences in total. On eight existing metonymy and metaphor classification benchmarks, augmenting training data with MetFuse consistently improves performance, with hybrid examples yielding the largest gains on metonymy tasks. Using this dataset, we also find that the presence of a metaphor makes a metonymic noun more explicit. Human annotators rate the metonymy in hybrid sentences as more explicit than in metonymy-only sentences (3.65 vs 3.47 out of 5). Models show the same pattern: augmenting BERT's training data with hybrid examples improves metonymy classification more than augmenting with metonymy-only examples across four benchmarks, and four LLMs more accurately identify metonymy in hybrid sentences than in
metonymy-only sentences in a zero-shot setting. In summary, our contributions are:
\begin{enumerate}[itemsep=2pt, topsep=4pt]
    \item \textbf{Framework:} We propose a framework to generate metonymy, metaphor, and hybrid variations of a literal expression.
    \item \textbf{MetFuse dataset:} We introduce MetFuse, the first dataset of figurative fusion between metonymy and metaphor, with 1,000 meaning-aligned quadruplets totaling 4,000 sentences.
    \item \textbf{Findings:} Our experimental results demonstrate that augmenting training data with MetFuse consistently 
    improves metonymy and metaphor classification across eight existing 
    benchmarks, and both humans and LLMs identify metonymy more easily when a 
    metaphoric verb is present.
\end{enumerate}

\section{Related Work}

\textbf{Metonymy \& Metaphor.} \citet{lakoff1980} introduced the Conceptual Metaphor Theory (CMT), the now-standard view that metaphor involves cross-domain mappings, which are relatively \textit{unconstrained and highly generative}~\citep{kovecses_2010_metaphor}. In contrast, metonymy operates within a single domain~\citep{nunberg_1995_transfer}, playing a crucial structural role in meaning construction~\citep{panther_2003_metonymy}, making it cognitively constrained~\citep{ruiz_2002_metonymy, ruiz_2005_metonymy, panther_2003_metonymy}, and therefore limited by pre-existing contiguity relations such as part--whole and container--content~\citep{radden, nerlich2001polysemy}. Unlike metaphor, metonymy is often approached as a cognitive and pragmatic phenomenon, rather than purely linguistic terms~\citep{radden, maria2015metonymy, papafragou1996metonymy}.

Computational approaches to metonymy have primarily focused on resolution tasks~\citep{markert-nissim-2007-semeval, gritta-etal-2017-vancouver, pedinotti-lenci-2020-dont, ghosh-jiang-2025-conmec}, treating the metonymy resolution as a classification task. Recently, metonymy has also been explored in a multimodal setting~\citep{ghosh-etal-2026-computational}. Metaphor identification has been an extensive field of research~\citep{Steen2010AMF, mohammad-etal-2016-metaphor, birke-sarkar-2007-active}. With the advent of LLMs, recent studies have focused on metaphor interpretation~\citep{chakrabarty-etal-2022-rocket, chakrabarty-etal-2022-flute}, probing LLMs' figurative language understanding capabilities. 

Existing literature has also studied metonymy and metaphor together. \citet{gossens} defined ``\textit{metaphtonymy},'' a phenomenon where metaphor and metonymy co-occur in figurative expressions. \citet{barcelona} shows how metaphor often builds on a metonymic base, or how metonymy becomes foregrounded under metaphoric pressure. \citet{maudslay-etal-2024-chainnet} introduces a structured resource that integrates metaphor and metonymy into WordNet, providing relational chains that capture figurative meaning extensions.

\begingroup
\renewcommand\baselinestretch{0.99}
\begin{table*}[t]
    \NiceMatrixOptions {
      custom-line = {
         command = dashedmidrule ,
         tikz = { dashed } ,
         total-width = \pgflinewidth + \aboverulesep + \belowrulesep ,
      } }
    \centering
    \small
    \begin{NiceTabular}{p{0.05\linewidth}|p{0.29\linewidth}|c|c|p{0.29\linewidth}}
    \toprule
    \multicolumn{1}{c}{\bf Noun} & \multicolumn{1}{c}{\bf Literal Input} & \multicolumn{1}{c}{\bf Candidates} &
    \multicolumn{1}{c}{\bf Scores} & \multicolumn{1}{c}{\bf Metonymic Sentence}  \\
    \midrule
    \multirow{3}{=}{judge} & \multirow{3}{=}{The \textsc{[mask]} in Lincoln saw issues} 
        & \textbf{law office} &  \textbf{-4.25} & \multirow{3}{=}{The \textbf{law office} in Lincoln saw issues}\\
        & & briefcase  & -12.28 & \\
        & & Springfield & -11.96 & \\
    \midrule
    \multirow{3}{=}{Queen} & \multirow{3}{=}{The \textsc{[mask]} officially pronounced Turing pardoned in August 2014.} 
        & crown & -1.58 & \multirow{3}{=}{\textbf{Buckingham Palace} officially pronounced Turing pardoned in August 2014.}\\
        & & monarchy & -11.58 & \\
        & & \textbf{Buckingham Palace} & \textbf{-0.27} & \\
    \bottomrule
    \end{NiceTabular}
    \caption{Illustrations of transforming literal expressions into metonymic ones using an LLM to generate candidate substitutions and masked token probabilities to select the best replacement.}
    \label{tab:metonymy_example}
\end{table*}
\endgroup

\noindent \textbf{Generation.} In the NLP space, early work in metaphor generation was heuristic in approach~\citep{abe_2006, terai_2010}. Recently, a lot of work has explored metaphor generation, including using literal-metaphor verb pairs~\citep{yu-wan-2019-avoid}, transforming literal expressions to metaphors~\citep{stowe-etal-2021-metaphor} and similes~\citep{chakrabarty-etal-2020-generating} using fine-tuned models. \citet{stowe-etal-2021-metaphor} proposed a supervised and unsupervised method to generate verbal metaphors, with the former yielding better results. The work of~\citet{stowe-etal-2021-exploring} analyzed verbal metaphor generation, finding that controlled generation improves metaphoricity, while free generation tends to generate more fluent paraphrases. Our metaphor generation strategy is inspired by~\citet{chakrabarty-etal-2021-mermaid}, which proposed a supervised method to generate verbal metaphors by replacing the relevant verb in the literal expression, and~\citet{tian-etal-2021-hypogen-hyperbole} that uses a partly-supervised framework to generate hyperbole without exploring the figurative expression it creates such as metaphor and sarcasm. Metonymy generation has not received much attention~\citep{hernandez2017paraphrasingverbalmetonymycomputational}, and our research aims to combine metonymy and metaphor generation.

\section{Methodology}

In this section, we discuss our methodology of generating metonymic, metaphoric and hybrid sentences from literal expressions. Our model takes a literal text as input, and transforms it into three figurative expressions: metonymy, metaphor and a hybrid sentence combining metaphor and metonymy.

\subsection{Preprocessing Steps}

Not all literal expressions can be transformed to its respective metonymy or metaphor counterpart. Metonymy occurs through a reference shift in the meaning of the noun, therefore, it is crucial that the literal sentence must contain a noun. Prior linguistic studies have shown that metonymy is more constrained than metaphor, since the mapping must remain within a single conceptual domain~\citep{gossens, kovecses_2010_metaphor}. In a canonical \textit{subject–verb–object} structure, both the \textit{subject} and the \textit{object} can in principle serve as sites of metonymic substitution. However, we restrict our focus to the subject noun for three reasons: (i) the \textit{subject} more often denotes animate, human entities, which are particularly productive for metonymic shifts, (ii) the \textit{subject} position offers more flexibility in shifting reference without disrupting grammatical structure, whereas \textit{object} substitutions can more easily lead to semantic or syntactic anomalies, and (iii) focusing on \textit{subjects} provides a consistent and controlled framework for generation, while still capturing the core phenomenon. Accordingly, we impose three conditions: the noun must denote a human entity, it must function as the \textit{subject}, and it must be in a dependency relation with a verb (enabling us to study its interaction with verbal metaphors). Under these constraints, the resulting instances mainly correspond to location metonymy. We then use the SpaCy dependency parser to extract the literal sentences from Wikipedia dumps\footnote{{https://huggingface.co/datasets/wikimedia/wikipedia}} based on the conditions set above. 


\subsection{Generating Metonymy}

Generating metonymy is a challenging task. The noun undergoes a reference shift, but the mapping must be intra-domain. Our framework contains three different stages---i) using an LLM to generate candidate replacement nouns, ii) identifying the best candidate noun, iii) replacing the noun and asking an LLM to refine the sentence.\\
\textbf{i. Generating candidate metonymic nouns.} We first provide the LLM with the literal sentence and the target noun. Naively asking it to ``make a metonymic replacement'' yields incorrect responses. Instead, we draw out the contiguity relations that evoke metonymy. We pose short, targeted questions about the noun's location, occupants/constituents, or salient parts. Each answer is treated as a candidate replacement for the noun phrase. We set a temperature of 0.7 to encourage diversity.\\
\textbf{ii. Selecting the metonymic noun.} To select the most suitable metonymic candidate from those generated, we used a masked language modeling approach with BERT. Specifically, we replace the target noun in the literal sentence with a \textsc{[mask]} token and present the sentence to BERT. For each candidate \textit{c}, we compute its log probability of filling the masked position. The candidate with the highest probability is chosen as the replacement, as it is deemed the most contextually appropriate. Table~\ref{tab:metonymy_example} shows an example. \\
\textbf{iii. Replacing the noun.} We select candidate $c*$ with the highest probability and substitute it for the original noun, yielding the metonymic variant of the sentence. To address any syntactic or grammatical inconsistencies introduced by the substitution, we then prompt the LLM to refine the sentence while preserving the metonymic noun. We use a temperature of 0.4 to discourage creative rewrites, ensuring that the refinement maintains high semantic similarity with the original literal sentence.

\subsection{Generating Metaphor}

In this work, we focus on generating verbal metaphors by mapping the verb to another domain. While metonymy typically arises through a reference shift in the noun, the verb also plays a crucial role in meaning construction~\citep{radden}. Mapping the verb to another domain makes the metaphor and metonymy directly dependent, enabling us to study the relation between them. Importantly, linguistic studies highlight that metaphor mapping allows for greater structural freedom than metonymy, which is typically constrained by contiguity relations~\citep{kovecses_2010_metaphor}. Building on this observation, \citet{stowe-etal-2021-exploring} showed that in verbal metaphor generation, controlled settings tend to increase metaphoricity, whereas free generation produces more fluent paraphrases. Motivated by these findings, and given the expressive power of recent LLMs, we allow the model absolute freedom in selecting the metaphoric domain—unlike in the case of metonymy—thereby exploiting this structural flexibility. To further intensify the domain shift, we also incorporate hyperbolic verbs in the spirit of \citet{tian-etal-2021-hypogen-hyperbole}, since hyperboles often make the metaphoric mapping more dramatic and salient.

Similar to metonymy generation, we use a 3-step pipeline---i) using an LLM to generate candidate verbal hyperboles, ii) identifying the best candidate verb, iii) using the verb and asking an LLM to refine the sentence.\\
\textbf{i. Generate metaphor candidates.} We provide the LLM with the literal sentence and the target verb. To generate verbal hyperboles, we instruct the model to exaggerate the verb to the extent it maps to another domain~\citep{lakoff1980}. However, we observed that these hyperboles often clashed with the overall tone of the sentence. Since \citet{lakoff1980, mohammad-etal-2016-metaphor} noted that tone plays a crucial role in metaphor interpretation, we incorporate this insight by asking the LLM to generate verbal hyperboles under three distinct tones---\textit{positive, negative} and \textit{neutral}, with a temperature of 0.7 and top-p 0.9 to encourage diversity. This additional context provides the LLM more flexibility and guidance, leading to coherent candidate generations.\\
\textbf{ii. Selecting the hyperbole verb.} To select the most suitable metaphor candidate verb that aligns with the overall tone of the literal sentence, we use a lightweight sentiment analysis model~\citep{camacho-collados-etal-2022-tweetnlp}. Based on the  predicted sentiment, we select the hyperbole candidate that best matches with the tone.\\
\textbf{iii. Refining the Metaphor.} We then replace the target word in the literal expression with the selected hyperbole candidate. We provide the sentence to the LLM and ask it to refine the sentence. We prompt it to maintain the metaphoric meaning while making any syntactical or structural adjustments to improve the sentence quality with a temperature of 0.6 and top-p 0.9.

\subsection{Constructing Hybrid Expressions}

To make the hybrid sentences, we take the metonymic noun phrase from the refined metonymic sentence, and replace it with the noun phrase in the refined metaphor sentence. While the metaphor generation step often alters the structure of the literal sentence—reflecting the greater freedom of metaphoric mapping—metonymy is more constrained in nature~\citep{radden}. As a result, metonymic generation typically preserves the original syntactic structure, modifying only the noun phrase. Consequently, inserting the metonymic noun phrase into the metaphor sentence produces a hybrid expression without the need for additional post-processing.

We extracted literal sentences from Wikipedia as mentioned earlier in this section, and fed them into our framework to generate the metonymic, metaphoric and hybrid variants of the sentence. For our main experiment, we use Llama-3.1-8B.

\section{Framework Evaluation \& Dataset}

In this section, we first evaluate our framework using both human and automatic evaluation. Then, we use this framework to create the MetFuse dataset, the first dataset to contain instances of metonymy and metaphor combined.

\subsection{Evaluation of Our Framework}

To compare against our proposed method, we employ a general-purpose prompting baseline. For each literal sentence, the LLM is queried in three independent passes to produce: (i) a metonymic variant: given the sentence and target noun, the LLM is asked to replace the noun with a metonymic paraphrase, (ii) a metaphoric variant: given the sentence and target verb, the LLM is asked to transform the verb into a verbal metaphor, and (iii) a hybrid variant combining both: given the sentence, noun, and verb, the LLM is asked to introduce a metonymic paraphrase complemented by a verbal metaphor. Each pass uses a carefully designed chain-of-thought prompt for that particular variant with few-shot exemplars, ensuring fair comparison.

\noindent \textbf{Human Evaluation.} We pick 250 literal sentences from our pool and use our framework to generate the three figurative variants. We also use the general prompting method to generate the three variants for the same  sample of 250 literal sentences. Human annotators are then asked to classify the sentences. A sentence is classified as positive if: (i) it carries the intended figurative expression, and (ii) original meaning of the sentence remains intact.

\begin{table}[t]
    \centering
    \resizebox{0.9\linewidth}{!}{
    \begin{tabular}{lccc}
    \toprule
     & \textbf{Metonymy} & \textbf{Metaphor} & \textbf{Hybrid}  \\
    \midrule
    General & 38.8\% & 70.8\% & 49.2\%  \\
    Ours & \textbf{75.2\%} & \textbf{84.0\%} & \textbf{74.0\%}  \\
    \bottomrule
    \end{tabular}}
    \caption{Percentage of sentences evaluated by humans to have the intended figurative expression in a sample of 250. General row are sentences from basic prompting. The sentences generated by our framework are significant better with figurative expressions.}
    \label{tab:human_eval}
\end{table}

\begin{table}[t]
    \centering
    \resizebox{0.9\linewidth}{!}{
    \begin{tabular}{lccc}
    \toprule
     & \textbf{Metonymy} & \textbf{Metaphor} & \textbf{Hybrid}  \\
    \midrule
    General & 0.70 & 0.60 & 0.44  \\
    Ours & \textbf{0.84} & \textbf{0.82} & \textbf{0.70}  \\
    \bottomrule
    \end{tabular}}
    \caption{Cosine similarity score between the original literal sentence and the generated variant using a sentence transformer. Our framework better preserves the semantic meaning of the sentence.}
    \label{table:auto_eval}
\end{table}

Table~\ref{tab:human_eval} shows the result of the human evaluation. Our framework consistently outperforms the general method in generating the figurative expressions from a literal text. LLMs particularly struggle with generating metonymic variations, as only 38.8\% of the sentences generated using a general method were judged as metonymy, compared to 75.2\% of the sentences generated by our framework. Overall, our framework has a consistently better performance, with 84.0\% of the sentences labeled as metaphors, and 74.0\% containing both metonymy and metaphor.

\begingroup
\renewcommand\baselinestretch{0.99}
\begin{table*}[t]
    \NiceMatrixOptions {
    custom-line = {
       command = dashedmidrule ,
       tikz = { dashed } ,
       total-width = \pgflinewidth + \aboverulesep + \belowrulesep ,
     } }
    \centering
    \small
    \resizebox{\textwidth}{!}{
    \begin{NiceTabular}{|p{0.25\linewidth}|p{0.25\linewidth}|p{0.25\linewidth}|p{0.25\linewidth}|}
    \toprule
    \multicolumn{1}{c}{\bf Literal Input} & \multicolumn{1}{c}{\bf Metonymy} &
    \multicolumn{1}{c}{\bf Metaphor} & \multicolumn{1}{c}{\bf Hybrid}  \\
    \midrule
    The researchers formed a rational statement of his question. & The \textbf{\tb{laboratory}} formed a rational statement of his question. & The researchers \textbf{\tg{sculpted}} a rational statement of his question.  & The \textbf{\tb{laboratory}} \textbf{\tg{sculpted}} a rational statement of his question.\\ 
    \midrule
    The reporter couldn't have done too good a job on you. & The \textbf{\tb{newsroom}} couldn't have done too good a job on you. & The reporter \textbf{\tg{butchered}} you in that interview. & The \textbf{\tb{newsroom}} \textbf{\tg{butchered}} you in that interview. \\ 
    \midrule
    The Queen officially pronounced Turing pardoned in august 2014. & \textbf{\tb{Buckingham Palace}} officially pronounced Turing pardoned in august 2014. & The Queen's pardon for Turing \textbf{\tg{thundered}} through history in August 2014. & \textbf{\tb{Buckingham Palace's}} pardon for Turing \textbf{\tg{thundered}} through history in August 2014. \\ 
    \midrule
    The police exposed the crime ring in 1956. & \textbf{\tb{NYPD}} exposed the crime ring in 1956. & The police \textbf{\tg{unearthed}} the crime ring in 1956. & \textbf{\tb{NYPD}} \textbf{\tg{unearthed}} the crime ring in 1956. \\ 
    \bottomrule
    
    \end{NiceTabular}
    }
    \caption{Example of a literal expression and its respective figurative variations from MetFuse dataset.}
    \label{tab:MetFuse_dataset_example}
\end{table*}
\endgroup

\noindent \textbf{Automatic Evaluation.} While the human evaluation showed the percentage of sentences judged to have the intended expression, an important aspect of transforming literal expressions to their figurative variations is that the generated sentences must be faithful to the input~\citep{chakrabarty-etal-2021-mermaid}. To evaluate this criteria, we use sentence transformer~\citep{reimers-gurevych-2019-sentence} to calculate the semantic similarity between the generated sentences and the original input sentence. Table~\ref{table:auto_eval} shows the result of this evaluation. Metonymic sentences have higher semantic similarity with the literal expression, followed by metaphor, while hybrid expressions have the least. The sentences generated by our framework show significantly higher semantic similarity than the general method. The sentences generated by general prompting (without any structural guidance) are semantically different from the literal text, with the LLM altering its structure, which often leads to loss or change in meaning. 

Overall, the human and automatic evaluation on 250 samples shows that naive prompting struggles to generate metonymic and hybrid sentences from literal texts. The semantic structure of the generated text also shifts to the degree of a loss or alteration of meaning. Our framework significantly outperforms the baseline, generating intended figurative variations of the literal text that are also semantically similar to its source. 

\subsection{MetFuse Dataset}

\begin{table}[t]
    \centering
    \resizebox{0.99\linewidth}{!}{
    \begin{tabular}{lccc}
    \toprule
     & \textbf{Metonymy} & \textbf{Metaphor} & \textbf{Hybrid} \\
    \midrule
    Fluency & 3.30 & \textbf{3.64} & 3.61  \\
    Meaning & \textbf{3.51} & 3.10 & 2.74 \\
    Creativity & 2.95 & 4.01 & \textbf{4.25} \\
    Metonymicity & 3.47 & - & \textbf{3.65} \\
    Metaphoricity & - & \textbf{3.95} & 3.82 \\
    \bottomrule
    \end{tabular}}
    \caption{Human score on five criteria from 250 samples from the MetFuse dataset.}
    \label{tab:human_ratings}
\end{table}

Leveraging our figurative-to-literal framework, we construct the MetFuse dataset. To build this dataset, human annotators reviewed 1,500 samples generated by our framework. Of these, 1,104 were judged to contain all three intended figurative variants, with around 74\% accuracy, as shown in Table~\ref{tab:human_eval}. From this pool, we randomly sampled 1,000 instances to form the final MetFuse dataset. The final dataset contains 1,000 literal sentences, each paired with its metonymic, metaphoric and hybrid variants, resulting in a total of 4,000 sentences.  The inter-annotators score between the annotators, measured as the raw agreement was measured to be 96.3\% for metonymy, 91.2\% for metaphor and 91.1\% for hybrid sentences. Table~\ref{tab:MetFuse_dataset_example} shows some illustrative examples from the dataset.

\noindent \textbf{Human Score.} The MetFuse dataset provides researchers the unique opportunity to study interaction and entanglement of metonymy and metaphor within a single expression. To this end, we recruit human annotators to rate the figurative expressions from the MetFuse dataset. Each annotator was provided with a sample and asked to rate the figurative texts on a scale of 1 to 5, with the literal sentences as references. We used four criteria from \citet{chakrabarty-etal-2021-mermaid}, in addition to one of ours: (1) \textit{Fluency} (``How fluent, grammatical, well formed and easy to understand are the generated utterances?''), (2) \textit{Meaning} (``Are the input and the output referring or meaning the same thing?'') (3) \textit{Creativity} (``How creative are the generated utterances?”), (4) \textit{Metonymicity} (``How explicit is the metonymy?'') and (5) \textit{Metaphoricity} (``How explicit is the metaphor?'').

Table~\ref{tab:human_ratings} shows the human scores. Metaphor sentences are rated to have the highest fluency. Metonymy is judged as less creative and fluent, but they best conserve the meaning. The fluency, meaning, and creativity ratings corroborate with previous linguistic works that states that metonymy is structurally limited and play a crucial role in meaning preservation~\citep{radden, ruiz_2005_metonymy}, while metaphor is more nonrestrictive~\citep{kovecses_2010_metaphor}, leading to higher creativity and lower meaning preservation. Hybrids have the highest creativity score due to both the noun and verb being altered in a figurative way. But this tends to lose the implicit meaning, having the lowest meaning score. Table~\ref{tab:human_ratings} also shows that metaphor sentences were judged to have higher metaphoricity (how explicit is the metaphor), slightly edging out hybrid sentences. Interestingly, hybrid sentences were rated as having noticeably higher metonymicity (how explicit is the metonymy) over metonymic sentences. The results indicates that the human judges found the metonymy to be more explicit in the sentences when it was accompanied by a verbal metaphor.

\section{Analysis}
\label{experiment_results_section}

In this section, we analyze the MetFuse dataset through an extrinsic evaluation, examining how metaphoric verbs influence metonymic noun.

\subsection{Extrinsic Evaluation}

While the MetFuse dataset is helpful for analyzing how metonymy and metaphor interact together, it can also be used for other tasks. To showcase this, we perform an extrinsic evaluation. Specifically, we examine if our dataset can improve metaphor and metonymy classification using existing datasets via data augmentation.

\noindent \textbf{Datasets.} We use four existing metonymy datasets: two common noun metonymy datasets---ConMeC~\citep{ghosh-jiang-2025-conmec} and \citet{pedinotti-lenci-2020-dont}, and two named entity datasets---RelocaR~\citep{gritta-etal-2017-vancouver} and WiMCor~\citep{alex-mathews-strube-2020-large}. For metaphor, we use four verbal datasets as well: VUA Verb~\citep{Steen2010AMF}, Flute~\citep{chakrabarty-etal-2022-flute}, MOH-X~\citep{mohammad-etal-2016-metaphor}, and TroFi~\citep{birke-sarkar-2007-active}. All eight datasets are binary classification tasks---a system should determine whether the given sentence contains a figure of speech or not.

\noindent \textbf{Experimental Setup.} For a given dataset, we use a 70–30 train–test split and fine-tune BERT-base~\citep{devlin-etal-2019-bert} under three settings: (i) the original training samples (Train), (ii) the training set augmented with MetFuse figurative examples (Train + \(\text{MTF}_{mty}\), or Train + \(\text{MTF}_{mtr}\)), and (iii) the training set augmented with MetFuse hybrid examples. For the metonymy classification task, the figurative examples in (ii) are metonymic variants, whereas for the metaphor downstream task, the same setup is followed except that the added samples are metaphoric variants. In all cases, the MetFuse augmentation size is fixed at 50\% of the original training set. We fine-tune BERT for 3 epochs with a learning rate 1e-5 and a batch size 8.

\begin{table}[t]
    \centering
    \resizebox{0.98\linewidth}{!}{
    \begin{tabular}{lccc}
    \toprule
     & \textbf{Train} & \textbf{\makecell{Train\\+\(\text{MTF}_{mty}\)}} & \textbf{\makecell{Train\\+\(\text{MTF}_{hyb}\)}} \\
    \midrule
    ConMeC & 75.49 & 76.71 \tg{{\scriptsize (+1.22)}} & \textbf{79.33} \tg{{\scriptsize (+3.84)}} \\
    Pedinotti & 68.42 & 66.92 \tr{{\scriptsize (-1.50)}} & \textbf{70.44} \tg{{\scriptsize (+2.02)}} \\
    RelocaR & 67.33 & 69.99 \tg{{\scriptsize (+2.66)}} & \textbf{70.67} \tg{{\scriptsize (+3.34)}} \\
    WiMCor & 81.67 & 82.33 \tg{{\scriptsize (+0.66)}} & \textbf{82.67} \tg{{\scriptsize (+1.00)}} \\
    \bottomrule
    \end{tabular}}
    \caption{Test accuracy for downstream tasks on metonymy datasets. Each dataset is split in a 70-30 train-test ratio. Train = original training sample, Train+\(\text{MTF}_{mty}\) = original training sample augmented with metonymic examples from MetFuse, Train+\(\text{MTF}_{hyb}\) = original training sample augmented with hybrid examples from MetFuse.}
    \label{tab:add_metonymy_downstream}
\end{table}

\noindent \textbf{Metonymy results.} Table~\ref{tab:add_metonymy_downstream} presents the results of the downstream experiment on metonymy datasets. Test accuracy improves when the training set is augmented with metonymic samples from MetFuse in three out of the four datasets, with only~\citet{pedinotti-lenci-2020-dont} being the exception. Notably, across all datasets, the highest test accuracy is achieved when the training set is augmented with hybrid examples. This suggests that the model learns the metonymic usage of a noun more effectively when exposed to a co-occurring metaphoric verb. This pattern aligns with the findings in Table~\ref{tab:human_eval}, where human annotators also found metonymic nouns easier to identify when paired with metaphoric verbs.

\noindent \textbf{Metaphor.} Table~\ref{tab:met_downstream} presents the results of the metaphor downstream experiment. Test accuracy improves consistently across all four datasets when the original training data is augmented with the MetFuse samples, underscoring the usefulness and generalizability of MetFuse. Hybrid examples yield the best performance on VUA Verb and MOH-X, whereas metaphoric examples perform better on Flute and TroFi. This suggests that the influence of metonymy on metaphor is not uniform---unlike the more consistent effect of metaphor on metonymy. In some cases, hybrid constructions dominate, while in others, purely metaphoric examples are more effective.

\begin{table}
    \centering
    \resizebox{0.98\linewidth}{!}{
    \begin{tabular}{lccc}
    \toprule
     & \textbf{Train} & \textbf{\makecell{Train\\+\(\text{MTF}_{mtr}\)}} & \textbf{\makecell{Train\\+\(\text{MTF}_{hyb}\)}} \\
    \midrule
    VUA Verb & 64.38 & 64.53 \tg{{\scriptsize (+0.15)}} & \textbf{66.55} \tg{{\scriptsize (+2.17)}} \\
    Flute & 76.26 & \textbf{79.73} \tg{{\scriptsize (+3.47)}} & 78.47 \tg{{\scriptsize (+2.21)}} \\
    MOH-X & 76.26 & 76.92 \tg{{\scriptsize (+0.16)}} & \textbf{77.43} \tg{{\scriptsize (+0.51)}} \\
    TroFi & 60.42 & \textbf{63.30} \tg{{\scriptsize (+2.88)}} & 61.76 \tg{{\scriptsize (+1.34)}} \\
    \bottomrule
    \end{tabular}}
    \caption{Test accuracy for downstream tasks on metaphor datasets. Each dataset is split in a 70-30 train-test ratio. Train = original training sample, Train+\(\text{MTF}_{mty}\) = original training sample augmented with metaphor examples from MetFuse, Train+\(\text{MTF}_{hyb}\) = original training sample augmented with hybrid examples from MetFuse.}
    \label{tab:met_downstream}
\end{table}

\subsection{Metaphor Improves Metonymy}
\label{section5_2}

Our previous results highlighted that both humans (Table~\ref{tab:human_ratings}) and supervised BERT (Table~\ref{tab:add_metonymy_downstream}) found it easier to identify the metonymic usage of a noun in hybrid sentences, i.e., when the metonymic noun was paired with a metaphoric verb. We investigate this further by performing metonymy resolution task using LLMs. We use two setups designed to test how the presence of metaphor influences metonymy identification: (i) treating metonymic sentences as positive cases and their literal counterparts as negative, and (ii) treating hybrid sentences as positive cases and their literal counterparts as negative. For each input, we provide the sentence and the noun to the LLM and ask if the noun is used metonymically~\citep{ghosh-jiang-2025-conmec}. We use four state-of-the-art models: GPT-OSS-20B~\citep{openai2025gptoss120bgptoss20bmodel}, Qwen3-30B~\citep{yang2025qwen3technicalreport}, Llama-3.1-70B~\citep{grattafiori2024llama3herdmodels}, and Gemini-2.5-Flash~\citep{comanici2025gemini25pushingfrontier}.

Table~\ref{table:main_table} shows the results. Across all four models, the precision, recall and F1 scores are consistently higher for hybrid samples than for purely metonymic ones. This suggests that LLMs can easily identify the metonymic usage of the noun when it co-occurs with a metaphoric verb.

\begin{table}[t]
    \centering
    \resizebox{0.99\linewidth}{!}{
    \begin{tabular}{lcccccc}
    \toprule
    \multirow{3}{*}{\textbf{Model}} & \multicolumn{3}{c}{\bf Metonymy} & \multicolumn{3}{c}{\bf Hybrid} \\
    \cmidrule(lr){2-4} \cmidrule(lr){5-7}  
    & Pre & Rec & F1 & Pre & Rec & F1 \\
    \midrule
    GPT-OSS-20B & 95.8 & 51.9 & 67.3 & \textbf{96.1} & \textbf{57.0} & \textbf{71.6} \\
    Qwen3-30B & 79.3 & 92.5 & 85.4 & \textbf{79.9} & \textbf{96.3} & \textbf{87.3} \\
    Llama-3.1-70B & 85.6 & 95.8 & 90.4 & \textbf{85.9} & \textbf{97.5} & \textbf{91.3} \\
    Gemini-2.5 & 92.4 & 95.4 & 93.9 & \textbf{92.5} & \textbf{97.0} & \textbf{94.7} \\
    \bottomrule
    \end{tabular}
    }
\caption{F1 score of positive metonymic sentences under two conditions: using metonymy only sentences as positive sentences, and using hybrid sentences as positive sentences. Literal sentences are the negative sentences.}
\label{table:main_table}
\end{table}

We investigate further with the token embeddings to check if similar patterns persist. We take the contextual embeddings of the nouns in the literal sentence (\(\text{N}_{lit}\)) and embeddings of the metonymic nouns in the metonymic sentences (\(\text{N}_{mty}\)) and calculate the cosine similarity between them. This effectively tells us how similarly the nouns are used in the literal and metonymic sentences. Similarly, we find the embeddings of the metonymic nouns in the hybrid sentences (\(\text{N}_{hyb}\)) and calculate the similarity with (\(\text{N}_{lit}\)). This tells us how similarly the nouns are used in the literal and hybrid sentences. We then compare \(sim(\text{N}_{lit},\text{N}_{mty})\) with \(sim(\text{N}_{lit},\text{N}_{hyb})\).

\begin{table}[t]
    \centering
    \resizebox{0.99\linewidth}{!}{
    \begin{tabular}{lcc}
    \toprule
     & \(sim(\text{N}_{lit},\text{N}_{mty})\) & \(sim(\text{N}_{lit},\text{N}_{hyb})\) \\
    \midrule
    GPT-OSS-20B & 71.00 & \textbf{72.38} \tg{{\scriptsize (+1.38)}}  \\
    Qwen3-30B & 90.88 & \textbf{91.75} \tg{{\scriptsize (+0.87)}}  \\
    Llama-3.1-70B & 57.78 & \textbf{59.64}  \tg{{\scriptsize (+1.86)}}\\
    BERT & 65.42 & \textbf{65.62} \tg{{\scriptsize (+0.20)}}\\
    \bottomrule
    \end{tabular}}
    \caption{Similarity score between the contextual embeddings of the noun tokens. \(sim(\text{N}_{lit},\text{N}_{mty})\) = similarity between the noun in the literal sentence and metonymic sentence. \(sim(\text{N}_{lit},\text{N}_{hyb})\) = similarity between the noun in the literal sentence and hybrid sentence.}
    \label{tab:cosine_noun}
\end{table}

\begingroup
\renewcommand\baselinestretch{0.95}
\begin{table*}[ht]
    \NiceMatrixOptions {
    custom-line = {
       command = dashedmidrule ,
       tikz = { dashed } ,
       total-width = \pgflinewidth + \aboverulesep + \belowrulesep ,
     } }
    \centering
    \small
    \resizebox{\textwidth}{!}{
    \begin{NiceTabular}{>{\centering\arraybackslash}p{0.08\linewidth}p{0.45\linewidth}p{0.45\linewidth}}
    \toprule
    \multicolumn{1}{c}{\bf Type} & \multicolumn{1}{c}{\bf Original Sentence} & \multicolumn{1}{c}{\bf Generated Sentence} \\
    \midrule
    \multirow{9}{*}{\textbf{Metonymy}} & (1) His \textbf{\tb{guitarist}} realizes what he did and knocks him out. & His \textbf{\tr{chord}} realizes what he did and knocks him out.\\ 
    \cmidrule(l){2-3}
    & (2) The reporters questioned, ``why would a \textbf{\tb{cricketer}} do this?'' & The reporters questioned, ``why would a \textbf{\tr{bat}} do this?''\\ 
    \cmidrule(l){2-3}
    & (3) His \textbf{\tb{father}} guided him in his early year. & His \textbf{\tr{wisdom}} guided him in his early years. \\
    \cmidrule(l){2-3}
    & (4) The \textbf{\tb{player}} was furious at the referee. & The \textbf{\tr{athlete}} was furious at the referee. \\
    \cmidrule(l){2-3}
    & (5) The \textbf{\tb{teacher}} encouraged her to apply for the position. & The \textbf{\tr{staff at the school}} encouraged her to apply for the position. \\
    \midrule
    \multirow{5}{*}{\textbf{Metaphor}} & (6) A dancer should \textbf{\tb{watch}} her diet carefully. & A dancer should \textbf{\tr{obsess}} over her diet carefully.\\
    \cmidrule(l){2-3}
    & (7) A great menacing student \textbf{\tb{warns}} her not to trust her family. & A great menacing student \textbf{\tr{poisoned her mind}}. \\
    \cmidrule(l){2-3}
    & (8) The economist \textbf{\tb{described}} the ``battle for the net''. & The economist \textbf{\tr{waged a war}}. \\
    \midrule
    \multirow{1}{*}{\textbf{Hybrid}} & (9) The \textbf{\tb{painters}} \textbf{\tb{worked}} on this masterpiece for 11 years. & The \textbf{\tr{studio}} \textbf{\tr{forged}} this masterpiece for 11 years.\\
    \bottomrule   
    \end{NiceTabular}
    }
    \caption{Error analysis of the generated sentences using our framework. Words highlighted in \textbf{\tb{blue}} in the original sentence are the target words that are being altered. The \textbf{\tr{red}} words are the cause of the errors in the generated sentence.}
    \label{tab:error_case}
\end{table*}
\endgroup

Table~\ref{tab:cosine_noun} shows the results. \(sim(\text{N}_{lit},\text{N}_{hyb})\) consistently has a higher value than \(sim(\text{N}_{lit},\text{N}_{mty})\). This means according to the LLM contextual embeddings, the same noun in the hybrid sentence is more similar to its literal sentence counterpart than in metonymic sentence. This suggests the metonymy in the hybrid sentence is more explicit as its embeddings are more similar to that of a non-metonymic usage. This corroborates with the human judgement in Table~\ref{tab:human_ratings}.

\noindent \textbf{Qualitative Discussion.} Our empirical results show that metonymy's strength tends to increase when paired with a metaphor. We look at this from a purely linguistic and cognitive point of view. In a purely metonymic sentence such as ``The \textit{newsroom} was harsh on the actor,' the noun \textit{newsroom} can remain cognitively unresolved due to metonymy's single domain mapping nature. Readers may or may not consciously resolve it to \textit{journalists}, since both the literal (place) and metonymic (people inside) readings remain available. When a metaphoric verb is introduced, the interpretive dynamics change. In ``\textit{The newsroom butchered the actor,}'' a verb like ``\textit{butchered}'' belongs to a semantic domain of physical violence, carrying strong selectional preferences for an animate, agentive subject. Because \textit{newsroom} is not literally animate, the metaphor exerts pressure on the reader to resolve the metonymy.

In this way, the out-of-domain mapping introduced by the metaphor forces explicit metonymy resolution, making the metonymy more salient than it would be in isolation. Thus, metaphor functions as a forcing device: its cross-domain mapping imposes constraints that push the metonymic noun into an explicit, agentive reading. This explains why in hybrid cases, metonymy is often perceived as more prominent and harder to ignore than in purely metonymic sentences. As a comparison, further analysis in Appendix~\ref{metonymy_impacts_metaphor} shows that the co-occurrence of metonymy does not impact metaphor's performance consistently.

\noindent \textbf{Takeaway.} Our analysis shows that humans and LLMs agree on one thing: \textit{a metaphoric verb can strengthen the metonymic nouns strength} if they co-occur in a sentence with a dependency relation.

\subsection{Error Analysis}

Table~\ref{tab:error_case} shows some qualitative examples of the error cases during generating metonymic, metaphoric and hybrid expressions using our framework. Sentence (1) and (2) are examples of the major error cases in generating metonymy, caused by the semantic structure of the original sentence. These instances of nouns renders itself difficult to be transformed to a metonymic variant, often yielding no natural metonymic replacements. In sentence (3), the metonymic substitution (\textit{father $\rightarrow$ wisdom}) alters the meaning of the original sentence. Sentence (4) is an instance of a literal substitution, an error made by Llama when generating metonymic substitutions. Sentence (5) is an example of the error occurring due to LLM paraphrasing after substituting the noun. In this case, Llama correctly generated a metonymic noun substitution (\textit{teacher $\rightarrow$ school}). However, when the LLM was asked to refine the already metonymic sentence ``\textit{The school encouraged her to apply for the position},'' the paraphrasing added the clause \textit{staff at the school}, rendering the sentence non-metonymic.

In sentence (6), the annotators agreed the verb is non-metaphoric. Sentence (7) and (8) are the major error cases in metaphor generation, with the meaning of the paraphrased sentence being inherently different from the literal source. In sentence (9), replacing the noun and the verb makes the hybrid expression creative, but the annotators agree that the sentence loses some of the intended meaning as the literal one. The literal expression refers to the art of painting, but one cannot understand the context just by looking at the hybrid expression (it can refer to any type of art, such as painting, singing, music, or sculpture).

\section{Conclusion}

We introduced a framework that transforms a literal sentence into
metonymic, metaphoric, and hybrid variants while preserving the 
meaning, and used it to construct MetFuse, the first dedicated 
dataset of figurative fusion between metonymy and metaphor. MetFuse
contains 1,000 meaning-aligned quadruplets totaling 4,000 sentences.
Across eight existing benchmarks, augmenting training
data with MetFuse consistently improves both metonymy and metaphor
classification. Our analysis further shows that both humans and large
language models identify metonymy more easily when a metaphoric verb is
present, suggesting that the metaphor's cross-domain mapping forces a
more explicit reading of the co-occurring metonymic noun. We hope
MetFuse will enable further study of how metonymy and metaphor interact
in context.


\section*{Limitations}

While our study makes important progress in addressing metonymy and metaphor interaction within NLP, certain limitations remain.

Our current work does not encompass the full range of metonymy that occurs in natural language. We limit ourselves to the animate subject nouns leading to location-for-people or institution-for-people metonymy, focusing on instances that are common and frequent. This choice allows us to build a strong foundation while avoiding excessive fragmentation of the problem space. We leave other metonymy types for future exploration. 
    
In our experiments, we do not explicitly compute or annotate the domain mappings for both metonymy and metaphor. Our goal is to lay the groundwork for metonymy-metaphor interaction study without constraining them to a fixed inventory of conceptual domain. We hope our work inspires future research to computationally study the domain mapping of these linguistic phenomena.

\section*{Ethical Considerations}

In this work, we employed large language models (LLMs) to generate candidate nouns and verbs for constructing metonymic, metaphoric, and hybrid expressions. For named-entity metonymy, in particular, the LLM was prompted to suggest location-based entities (e.g., cities, institutions) as replacements for target nouns. While this procedure leverages the generative capacity of LLMs, we recognize that such models may reproduce unintended biases or stereotypes, especially when dealing with named entities (e.g., associating certain locations with actions). We did not observe such issues in our generated samples; however, we acknowledge that these risks are possible. Importantly, our dataset does not include any private or personally identifiable information: all named entities are in the public domain. By acknowledging these limitations and safeguards, we emphasize that our use of LLMs is confined to controlled generation for research purposes, with human oversight applied to ensure that the final dataset avoids harmful content.

\section*{Acknowledgments}

We sincerely thank the CincyNLP group for their valuable feedback and help in annotation. We also thank the anonymous ARR reviewers for
their insightful suggestions and discussions.

\bibliography{main}

\newpage

\appendix

\section{Generalization of our Framework}

To assess how well our framework generalizes across different
LLMs, we use human to annotate 100 quadruplets generated by different models. Table~\ref{tab:generalization} shows the results, which highlights that our framework generalizes well across models of all sizes. The success rate of the framework depends primarily on the semantic structure of the input sentence, and not on the LLMs capability. Due to this, we chose Llama 3.1-8B to as the base model in our framework, as it keeps our model lightweight without compromising performance.

\begin{table}[h]
    \centering
    \resizebox{0.98\linewidth}{!}{
    \begin{tabular}{lccc}
    \toprule
     & \textbf{Metonymy} & \textbf{Metaphor} & \textbf{Hybrid} \\
    \midrule
    Llama-3.1-8B & 74 & 86 & 73 \\
    GPT-oss-20B & 74 & 87 & 74 \\
    Qwen3-30B & \textbf{75} & 86 & \textbf{75} \\
    Llama-3.1-70B & \textbf{75} & 87 & 74 \\
    GPT-5 & 72 & \textbf{90} & 72 \\
    \bottomrule
    \end{tabular}}
    \caption{Number of sentences evaluated by humans to have the intended figurative expression in a sample of 100 when ran on our framework with different LLMs.}
    \label{tab:generalization}
\end{table}

\subsection{Diversity}

An important factor in generating the metonymic replacements of the noun is diversity. Open-source LLMs like Llama and Qwen are more suited to our framework as it gives the user flexibility in terms of diversity by increasing temperature and top-p value. While being the latest model in the bunch, GPT-5 suffered from repetitive replacements. We ran experiments with different ``thinking effort'' conditions, but the repetitiveness did not subside.

\section{Metonymy's Impact on Metaphor}
\label{metonymy_impacts_metaphor}

To analyze the impact of metonymy on metaphors, we repeat the token embedding similarity experiment conducted in Section~\ref{section5_2}, but this time, we compare the embeddings of the verb. Specifically, we find how similar the verbs are in the literal and metaphoric sentence. We calculate the similarity between the verb in the literal sentence and the metaphoric sentence \(sim(\text{V}_{lit},\text{V}_{mtr})\), and compare it with the verb in the literal sentence and hybrid sentence \(sim(\text{V}_{lit},\text{V}_{hyb})\). 

\begin{table}[ht]
    \centering
    \resizebox{0.99\linewidth}{!}{
    \begin{tabular}{lcc}
    \toprule
     & \(sim(\text{V}_{lit},\text{V}_{mtr})\) & \(sim(\text{V}_{lit},\text{V}_{hyb})\) \\
    \midrule
    GPT-oss-20B & 70.19 & \textbf{70.28} \tg{{\scriptsize (+0.09)}}  \\
    Qwen3-30B & 92.83 & \textbf{93.07} \tg{{\scriptsize (+0.24)}}\\
    Llama-3.1-70B & \textbf{57.32} & 52.88 \tr{{\scriptsize (-4.44)}} \\
    BERT & \textbf{65.24} & 65.02 \tr{{\scriptsize (-0.22)}}\\
    \bottomrule
    \end{tabular}}
    \caption{Similarity score between the contextual embeddings of the verb tokens. \(sim(\text{V}_{lit},\text{V}_{mtr})\) = similarity between the verb and in the literal sentence and metaphoric sentence. \(sim(\text{N}_{lit},\text{N}_{hyb})\) = similarity between the verb and in the literal sentence and hybrid sentence.}
    \label{tab:cosine_verb}
\end{table}

Table~\ref{tab:cosine_verb} shows the results. The results are less consistent than Table~\ref{tab:cosine_noun}. GPT-oss-20B and Qwen3-30B shows higher similarity between the verb in the literal-hybrid pair, while Llama 3.1-70B and BERT shows higher similarity in literal-metaphor pair. This observation aligns with the metaphor downstream experiment results in Table~\ref{tab:met_downstream}. The impact of metonymic noun on the metaphor verb is not consistent. We believe there are deeper semantic complexities at play, and we leave this for future work.

\begin{table}[h]
    \centering
    \resizebox{0.95\linewidth}{!}{
    \begin{tabular}{lcccc}
    \toprule
    \multirow{2}{*}{\textbf{Model}} & \multicolumn{2}{c}{\bf Similarity} & \multicolumn{2}{c}{\bf Entailment} \\
    \cmidrule(lr){2-3} \cmidrule(lr){4-5} 
    & Mtr & Hyb & Mtr & Hyb\\
    \midrule
    GPT-OSS-20B & \textbf{86.45} & 84.82 & \textbf{96.03} & 93.84 \\
    Qwen3-30B & \textbf{84.79} & 83.34 & \textbf{94.54} & 94.31 \\
    Llama-3.1-70B & \textbf{82.02}& 78.59 & \textbf{85.94} & 79.40 \\
    Gemini-2.5 & \textbf{86.31} & 82.13 & \textbf{93.24} & 90.96 \\
    \bottomrule
    \end{tabular}
    }
\caption{Semantic similarity and entailment scores of metaphor (Mtr) and hybrid (Hyb) sentences with respect to their literal counterparts.}
\label{table:metaphor_main_table}
\end{table}

We also investigate how the presence of a metonymic noun impacts the  interpretability of the metaphor, i.e., if LLMs can recover the literal meaning from the metaphor expression~\citep{chakrabarty-etal-2022-rocket, chakrabarty-etal-2022-flute}. For this task, we provide the metaphor and hybrid sentence to the LLM and ask it to paraphrase them to their literal meaning. We then evaluate the paraphrased outputs using entailment~\citep{liu2019roberta} and semantic similarity scores. Table~\ref{table:metaphor_main_table} shows the results. We find that metaphor-only sentences achieve equal or higher similarity and entailment scores compared to hybrid sentences. This suggests that an LLMs ability to interpret a metaphor may decrease when the metaphor is anchored to a metonymic noun.

\section{Additional Metonymy Downstream Results}
\label{additional_metonymy_downstream}

\begin{figure*}[t]
    \centering
    
    \begin{subfigure}[b]{0.24\linewidth}
        \centering
        \includegraphics[width=\linewidth]{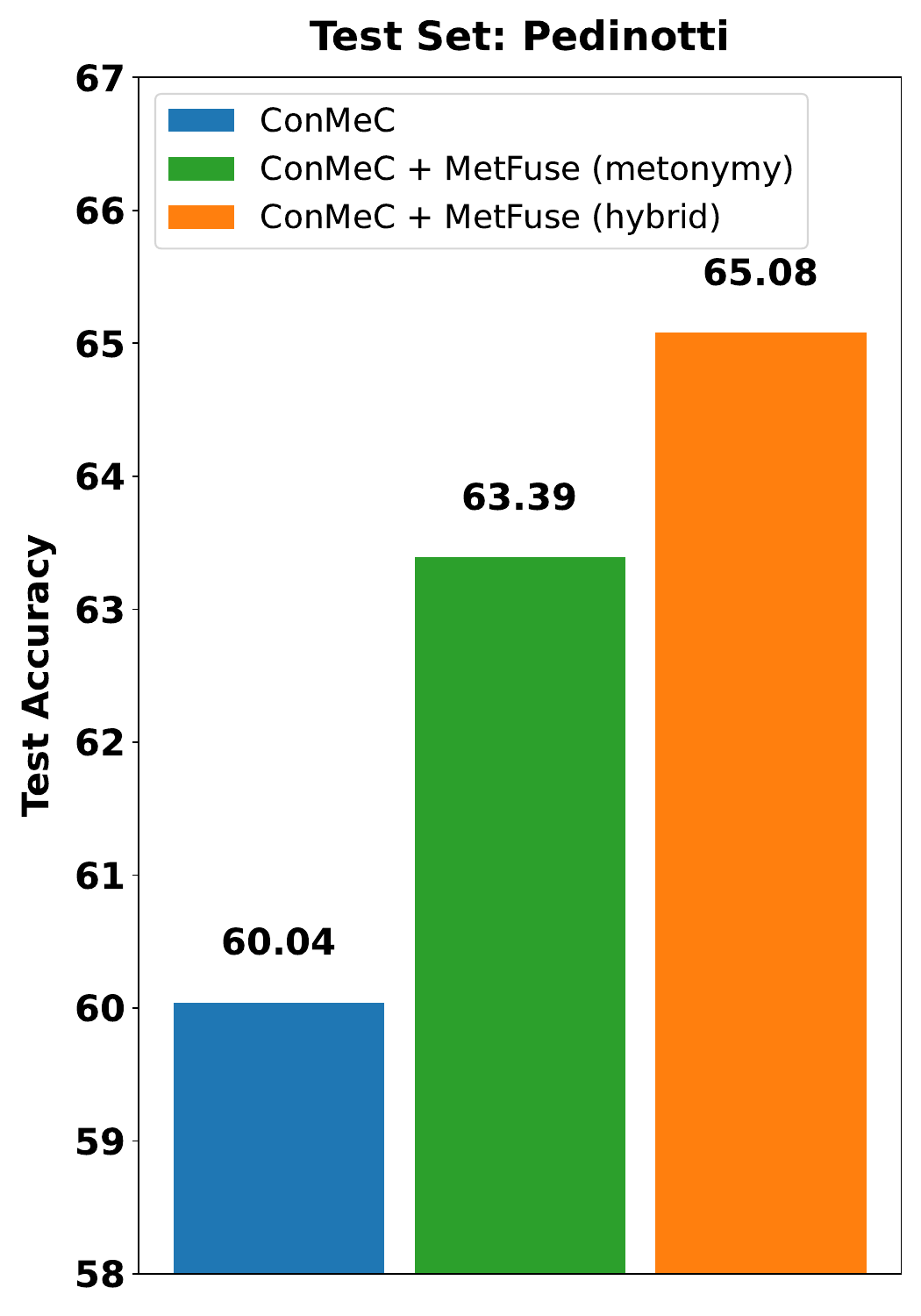}
        \caption{Train: ConMeC}
    \end{subfigure}
    \begin{subfigure}[b]{0.24\linewidth}
        \centering
        \includegraphics[width=\linewidth]{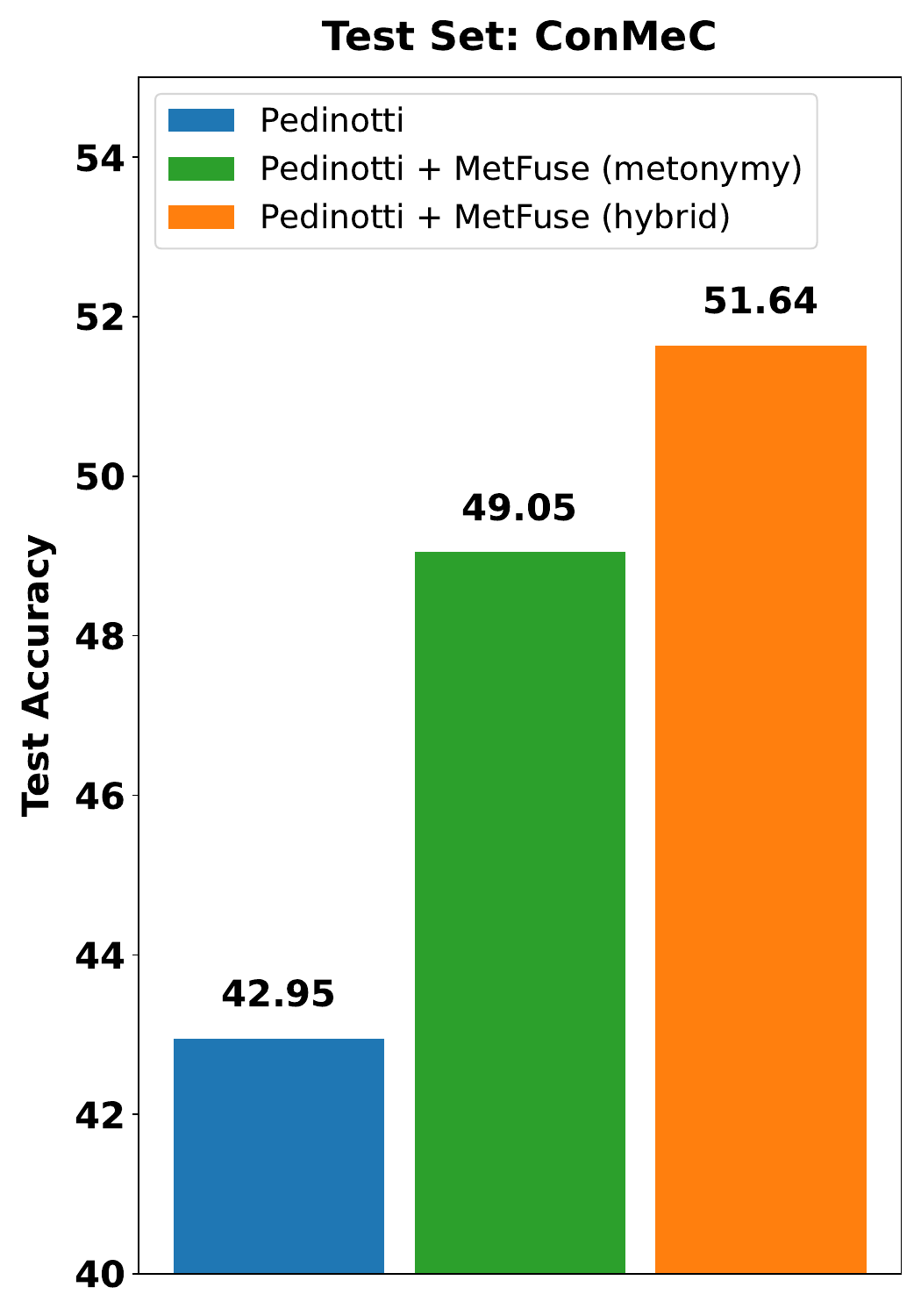}
        \caption{Train: Pedinotti}
    \end{subfigure}
    \begin{subfigure}[b]{0.24\linewidth}
        \centering
        \includegraphics[width=\linewidth]{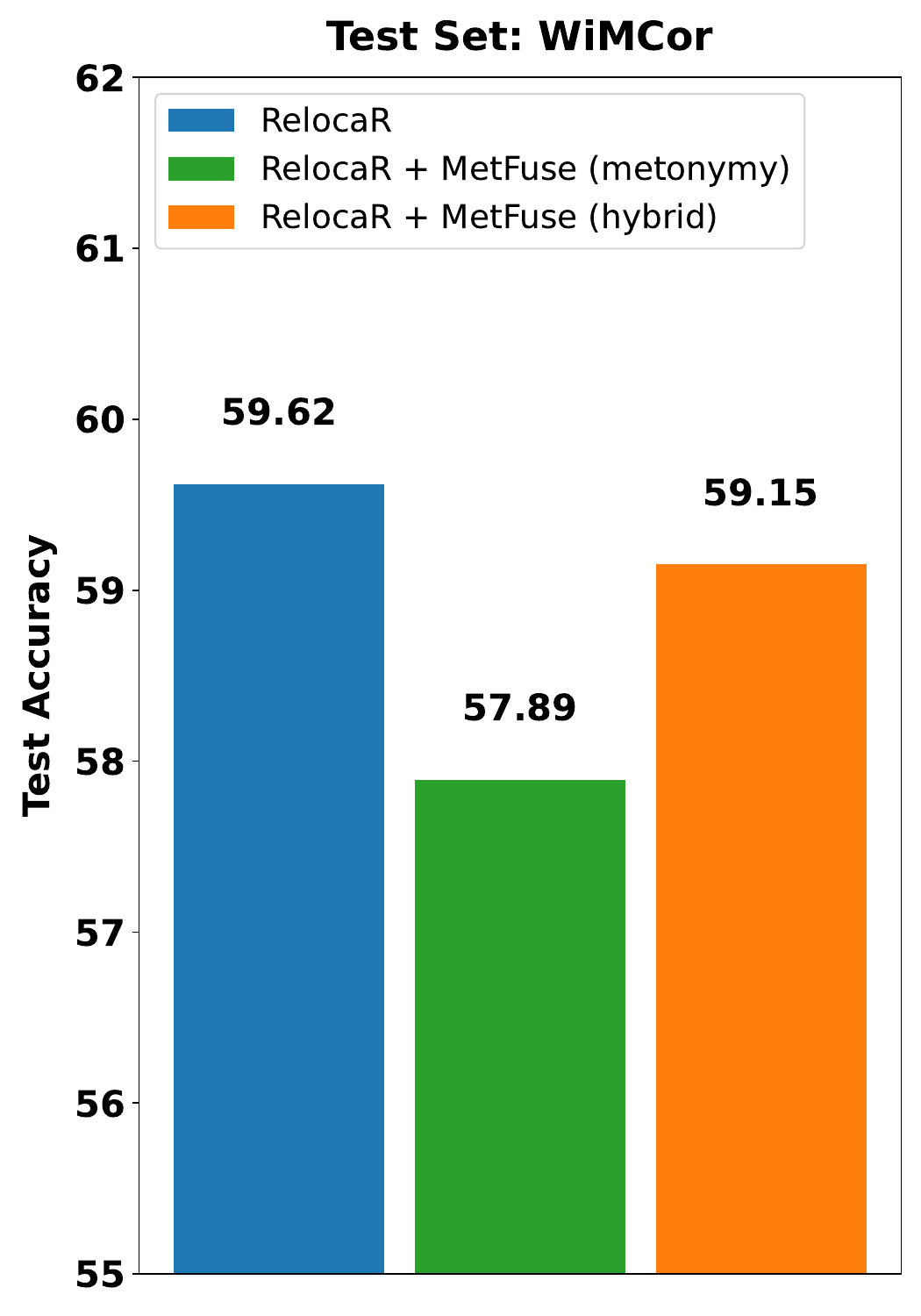}
        \caption{Train: RelocaR}
    \end{subfigure}
    \begin{subfigure}[b]{0.24\linewidth}
        \centering
        \includegraphics[width=\linewidth]{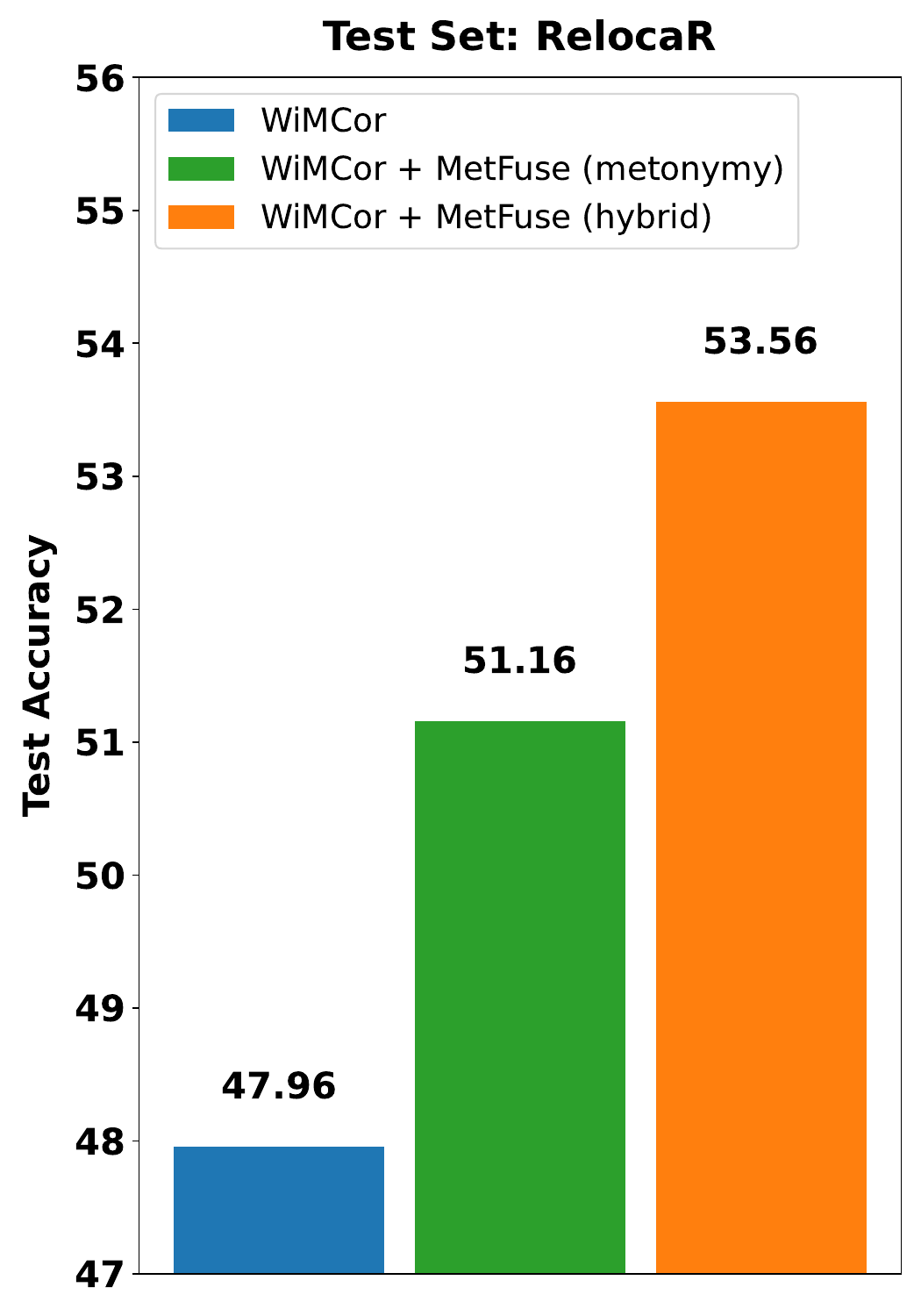}
        \caption{Train: WiMCor}
    \end{subfigure}
    
    \caption{Figure showing results of the downstream experiment. Blue bar shows the performance when trained on an existing metonymy dataset. Green bar shows the performance when trained on existing dataset + MetFuse metonymic samples, orange bar is existing datatset + MetFuse hybrid samples.}
    \label{fig:downstream}
\end{figure*}

Metonymy identification have been shown to struggle under cross-domain settings~\citep{ghosh-jiang-2025-conmec}. To this end, we employ the MetFuse dataset to study its impact in cross-domain metonymy classification. For this experiment, we train BERT on one dataset, and test it on another. We separate the two common noun datasets: ConMeC~\citep{ghosh-jiang-2025-conmec} and \citet{pedinotti-lenci-2020-dont}, and the two named entity datasets: RelocaR~\citep{gritta-etal-2017-vancouver} and WimCoR~\citep{nastase-strube-2009-combining}. When the model is trained on a common noun dataset, it is tested on a common noun dataset as well, and augmented with common noun examples from MetFuse. This goes the same for named entity dataset as well. This is done to ensure a fair setting, as MetFuse has both common noun and named entity examples. We keep the same conditions as before---the augmented MetFuse examples are 50\% the size of the original training sample, learning rate was set at 1e-5 over 3 epochs and batch size 8.

Figure~\ref{fig:downstream} shows the result of the cross-domain classification experiment. The test accuracy increases when the model is trained with the MetFuse dataset in all three cases, the only exception being when it is trained with RelocaR and tested on WimCoR. The hybrid examples from MetFuse always has a better performance than purely metonymic examples. The results highlight the usefulness and generalizability nature of the MetFuse dataset. It also shows the metonymic noun can be identified more easily by language models when it is accompanied by a metaphor, supporting the observation made in the main paper.

\section{Surprisal Scores}

For further analysis, we calculate the token surprisal scores from the MetFuse dataset. For a token $x$ with model probability $p(x)$, surprisal score is given by:
\[
s(x) \,=\, -\log p(x).
\]
Higher surprisal score indicates the model was not expecting this token, hence is ``\textit{surprised}''. For the literal, metonymy, metaphor and hybrid sentences, we calculate the surprisal of the noun token (responsible for metonymy) and the verb token (responsible for metaphor). 

\begin{table}[h]
    \centering
    \resizebox{0.95\linewidth}{!}{
    \begin{tabular}{lcccc}
    \toprule
    \textbf{Surprisal} & \textbf{Lit} & \textbf{Mty} & \textbf{Mtr} & \textbf{Hyb} \\
    \midrule
    Noun token & 9.01 & \textbf{12.79} & 9.65 & \textbf{12.81} \\
    Verb token & 7.03 & 9.02 & \textbf{11.38} & \textbf{12.66} \\
    \bottomrule
    \end{tabular}}
    \caption{Surprisal scores for the noun and verb token in literal (lit), metonymy (Mty), metaphor (Mtr) and hybrid (Hyb) sentences. Bold indicates the token was altered to create the intended figurative expression.}
    \label{tab:surprisal}
\end{table}

Table~\ref{tab:surprisal} shows the results of this experiment. The bold numbers indicate the instances where the word was changed by our framework to create the figurative expression. It is evident that the surprisal for the noun is high in metonymy and hybrid, while the surprisal for verb is high in metaphor and hybrid. 

\section{Discussion - Metonymy vs Metaphor Generation}

Our findings highlight an important asymmetry between metaphor and metonymy generation. Metonymy is significantly harder to generate in a controlled fashion. Its constraints come from the fact that metonymic substitutions are restricted to in-domain mappings~\citep{gossens}, which sharply narrows the candidate space. Moreover, the substituted expression must still refer to the same underlying entity, a requirement that is not always straightforward to satisfy For example, ``\textit{his father guided him}'' cannot simply be replaced by his father’s hand guided him without changing the referent).

In contrast, metaphor generation is comparatively more permissive~\citep{kovecses_2010_metaphor}. Because metaphors involve cross-domain mappings, a wider range of substitutions are tolerated, even in uncontrolled generation. While many generated metaphors may be novel or unconventional, they still tend to preserve intelligibility without the strict referential constraints that metonymy demands. 

These observations suggest that while metaphor generation can often succeed through broad lexical substitution, metonymy requires more fine-grained semantic control and discourse awareness. Evidently, table~\ref{tab:human_eval} shows 75.2\% of the sentences were judged to be metonymic by humans, compared to 84.0\% judged to be metaphors. A deeper investigation into when hybrids (metaphor–metonymy blends) outperform pure metaphors or pure metonymies is left for future work.

\end{document}